\documentclass[a4paper,11pt]{article}
\usepackage{graphicx}  
\usepackage{url}       
\usepackage{times}
\usepackage{natbib}
\usepackage[margin=25mm]{geometry}
\usepackage{subcaption}
\usepackage{placeins}
\usepackage{enumitem}
\usepackage{mwe}
\usepackage{multicol}
\DeclareCaptionFont{tiny}{\tiny}

\usepackage{amsmath,amsthm,amssymb,amsfonts,mathtools}

\usepackage{tabularx}
\usepackage{multirow}
\usepackage{wrapfig}

\usepackage[justification=justified,singlelinecheck=false]{caption}

\usepackage{algorithm}
\usepackage{algpseudocode}

\usepackage{tikz-dependency}
\usepackage{tikz}
\usetikzlibrary{positioning}
\definecolor {processblue}{cmyk}{0.96,0,0,0}

\newcommand{\APT}{%
	\textsc{Apt}}

\title{Temporal and Aspectual Entailment}
\date{}

\author{Thomas Kober\\
       University of Edinburgh\\
       \texttt{tkober@inf.ed.ac.uk}
  \and Sander Bijl de Vroe\\
       University of Edinburgh\\
       \texttt{sbdv@ed.ac.uk}
  \and Mark Steedman\\
       University of Edinburgh\\
       \texttt{steedman@inf.ed.ac.uk}
}

\begin{document}
\maketitle
\thispagestyle{empty}
\pagestyle{empty}

\begin{abstract}
Inferences regarding \emph{Jane's arrival in London} from predications such as \emph{Jane is going to London} or \emph{Jane has gone to London} depend on \emph{tense} and \emph{aspect} of the predications. Tense determines the temporal location of the predication in the past, present or future of the time of utterance. The aspectual auxiliaries on the other hand specify the internal constituency of the event, i.e. whether the event of \emph{going to London} is completed and whether its consequences hold at that time or not. 

While tense and aspect are among the most important factors for determining natural language inference, there has been very little work to show whether modern NLP models capture these semantic concepts. In this paper we propose a novel entailment dataset and analyse the ability of a range of recently proposed NLP models to perform inference on temporal predications. We show that the models encode a substantial amount of morphosyntactic information relating to tense and aspect, but fail to model inferences that require reasoning with these semantic properties.
\end{abstract}

\section{Introduction}
\label{introduction}
Tense and aspect are two of the main contributors to the semantics of a proposition, describing the temporal location of a predication and its internal constituency, thereby considerably influencing the entailment relations it licenses. For example, while \emph{arrive in LOC} $\models$ \emph{be in LOC} is generally considered a valid entailment rule, the case is complicated when different tenses and aspectual auxiliaries\footnote{For brevity we will refer to predications with different tenses and aspectual auxiliaries as \emph{temporal predications}.} of a given verb are considered as sentences~\ref{ex:has_arrived} and~\ref{ex:will_arrive} illustrate.
\begin{multicols}{2}
\begin{enumerate}[label={(\arabic*)},series=qa]
\itemsep0em
\item \label{ex:has_arrived} Jane \emph{has arrived} in London.
\item[] \hspace{0.5cm} $\models$ Jane \emph{is} in London now. 
\item \label{ex:will_arrive} Jane \emph{will arrive} in London. 
\item[] \hspace{0.5cm} $\not\models$ Jane \emph{is} in London now. 
\end{enumerate}
\end{multicols}
Understanding the difference between an event that has happened and whose consequences hold at the present moment, and an event that is currently happening or will happen in the future, is crucial for answering questions such as \emph{Where is Jane?} or \emph{Is Jane in London now?} Inferring the consequences of events is important for understanding the relation between entities in the world. For example, if we read that \emph{Lady Catherine has bought Longbourn estate}, the inference that the acquisition is \emph{completed}, and that the resulting consequence is that Lady Catherine now \emph{owns} Longbourn estate, is paramount for keeping knowledge bases up-to-date.

In this paper we propose a novel entailment dataset that requires models to correctly determine the internal and external temporal structure of predications when performing natural language inference. To the best of our knowledge, this is the first dataset that is primarily focused on assessing natural language inference between temporally and aspectually modified predications.

As a first evaluation on our new dataset we compare to what extent five distributional embedding models, \texttt{word2vec}~\citep{Mikolov_2013b}, Anchored Packed Trees~\citep{Weir_2016}, \texttt{fastText}~\citep{Bojanowski_2017}, ELMo~\citep{Peters_2018}, and BERT~\citep{Devlin_2018}, and two bi-directional LSTM (biLSTM) encoders, pre-trained on SNLI~\citep{Bowman_2015} and DNC~\citep{Poliak_2018}, respectively, are able to perform natural language inference on temporal predications. In our evaluation, we refrain from fine-tuning any of the models as our goal is to assess to what extent tense and aspect are captured in these models \emph{per se}. As a pre-requisite diagnostic task for natural language inference between temporal predications we analysed whether the models encode the morphosyntax of tense and aspect and found that they capture a considerable amount of morphosyntactic information in their respective embedding spaces. However, neither of the models outperforms a majority class baseline on our proposed dataset due to their reliance on contextual similarity for performing inference, suggesting that models based on distributional semantics struggle with the more latent nature of tense and aspect. Our contributions in this paper are as follows: 
\begin{itemize}
	\item We assess the extent to which the models in our evaluation encode information about the agreement between an inflected verb and its aspectual auxiliary, and whether a translation operation between different tenses can be learnt from the embedding spaces.
	\item We propose a novel entailment dataset that requires models to perform inference with temporal predications, and evaluate the five embedding models and two pre-trained biLSTM encoders.
	\item We analyse the performance of the models and show that their reliance on contextual similarity is problematic for correctly modelling natural language inference governed by tense and aspect.
\end{itemize}

\section{Tense, Aspect and Entailment}
\label{background}
Tense is a grammatical category which is encoded in the morphology of the verb in English (e.g. past \emph{loved} vs. non-past \emph{loves}). It establishes a point of reference that allows the temporal organisation of events in a discourse. In English, tense interacts with aspectual auxiliaries such as the verbs \emph{be} or \emph{have} that influence the internal constituency of a predication, and determine whether an event is completed or ongoing. Tense and aspect therefore control the internal and external temporal structure of an event and govern the inferences that a predication licenses~\citep{Reichenbach_1947,Dahl_1985,Steedman_1997}. There is evidence that such morphology is represented in distributional embeddings~\citep{Mitchell_2015,Vylomova_2016}. In this paper we are concerned with perfect and progressive aspect, but do not focus on any other types of aspect such as the \emph{Aktionsart} of a predication~\citep{Vendler_1957}, which we leave to future work.

\subsection{The Interaction between Temporality and Entailment}

Perfect aspect (typically) describes events as a completed whole, and licenses inferences regarding the consequences of that event. The use of different tenses and aspects for past events influences their relevance to the present moment and thereby their entailment behaviour. For example, the consequences of an event in the present perfect hold at the time of utterance, whereas events in the simple past or the past perfect do not~\citep{Comrie_1985,Moens_1988,Depraetere_1998,Katz_2003}. This is shown in sentences~\ref{ex:has_gone} and \ref{ex:went}, where only sentence~\ref{ex:has_gone} licenses the inference of Elizabeth being in Meryton \emph{now}.
\begin{multicols}{2}
\begin{enumerate}[label={(\arabic*)}, resume*=qa]
\itemsep0em
\item \label{ex:has_gone} Elizabeth \emph{has gone} to Meryton.
\item[] \hspace{0.5cm} $\models$ Elizabeth \emph{is} in Meryton now.
\item \label{ex:went} Elizabeth \emph{went} / \emph{had gone} to Meryton.
\item[] \hspace{0.5cm} $\not\models$ Elizabeth \emph{is} in Meryton now.
\end{enumerate}
\end{multicols}
This property can be explained through a Reichenbachian view of the present perfect, where the point of reference coincides with the point of speech, thereby indicating its current relevance~\citep{Reichenbach_1947}. On the other hand, events in the past simple or the past perfect license inferences for consequent states in the past, as sentence~\ref{ex:visited} shows.
\begin{multicols}{2}
\begin{enumerate}[label={(\arabic*)}, resume*=qa]
\itemsep0em
\item \label{ex:visited} Elizabeth \emph{went} / \emph{had gone} to Meryton.
\item[] \hspace{0.5cm} $\models$ Elizabeth \emph{was} in Meryton.
\item \label{ex:is_travelling} Mary \emph{is going} to Netherfield now.
\item[] \hspace{0.5cm} $\not\models$ Mary \emph{has arrived} / \emph{is} in Netherfield.
\end{enumerate}
\end{multicols}
Progressive aspect describes ongoing events and therefore does not license inferences regarding their consequences as sentence~\ref{ex:is_travelling} shows. It furthermore gives rise to the imperfective paradox~\citep{Dowty_1979}, which only seems to license inferences for non-culminated processes~\citep{Moens_1988}, as sentences~\ref{ex:was_walking} and~\ref{ex:was_reaching} show.
\begin{multicols}{2}
\begin{enumerate}[label={(\arabic*)}, resume*=qa]
\itemsep0em
\item \label{ex:was_walking} Catherine \emph{was walking} in the woods.
\item[] \hspace{0.5cm} $\models$ Catherine \emph{walked} in the woods.
\item \label{ex:was_reaching} Jane \emph{was reaching} London.
\item[] \hspace{0.5cm} $\not\models$ Jane \emph{reached} / \emph{was in} London.
\end{enumerate}
\end{multicols}
The modal future introduces an event whose realisation is uncertain, therefore any inferences about its outcome are only licensed if common-sense knowledge suggests that this is almost always the course of events as sentence~\ref{ex:will_meet} shows. 
\begin{enumerate}[label={(\arabic*)}, resume*=qa]
\itemsep0em
\item \label{ex:will_meet} Charles \emph{will meet} with Jane.
\item[] \hspace{0.5cm} $\models$ Charles \emph{will see} Jane.
\end{enumerate}

The correct treatment of tense and aspect in a predication is crucial for inferring the consequences it licenses, which is important for answering questions about a given paragraph, or creating and updating knowledge bases.

\section{Models}
\label{methodology}
We analyse five distributional embedding models and two pre-trained biLSTM sentence encoders for their ability to perform inference on temporal predications. Our choice of models is motivated by the observation that modelling entailment between temporal predications requires a bespoke representation of the inflected verb in the context of the given aspectual auxiliary and its arguments.

\textbf{\texttt{word2vec}.} We evaluate the ability of \texttt{word2vec} representations for performing inference with temporal predications. Contextualisation\footnote{We refer to expressing the meaning of a word in its context as \emph{contextualisation}.} can be achieved by averaging two word vectors, which has been shown to be a strong baseline for a range of problems~\citep{Iyyer_2015,Wieting_2016b}. Notably, adding or averaging word vectors approximates the intersection of their feature spaces~\citep{Tian_2017}.

\textbf{{\APT}s.} Anchored Packed Trees are a recently proposed vector space model that take distributional composition to be a process of lexeme contextualisation. {\APT}s are based on a higher-order dependency-typed structure that gives rise to a weighted, directed and labelled graph. Contextualisation is achieved through distributional composition, which requires aligning two lexemes according to their syntactic relation, and then merging the aligned representations. {\APT}s are the only count-based (i.e. non-neural) model in our evaluation.

\textbf{\texttt{fastText}.} The \texttt{fastText} model represents each word as a sum of bag-of-character n-grams, thereby making better use of subword information and therefore --- potentially --- providing a better mechanism for encoding morphosyntactic relations. Contextualisation is achieved through averaging the respective word vectors in a phrase.  

\textbf{ELMo.} ELMo is based on a deep bidirectional LSTM language model that creates multiple layers of representations for every token. Contextualised representations are obtained from the internal states of the LSTMs, where~\citet{Peters_2018} showed that lower levels of the architecture capture syntactic characteristics, and higher-levels capture semantic characteristics of words.

\textbf{BERT.} BERT uses multi-headed bi-directional self-attention and is based on the Transformer architecture~\citep{Vaswani_2017}. \citet{Devlin_2018} observed that sequential language model architectures are limited by the unidirectionality of the models. Therefore they proposed a novel training objective that jointly conditions on left and right context in all layers. They showed that their training regime results in substantial gains over serial language model-based architectures on numerous NLP tasks.

\textbf{Pre-trained biLSTM.} For our new entailment dataset, we pre-trained two bi-directional LSTM~\citep{Hochreiter_1997} sentence encoders on SNLI~\citep{Bowman_2015} and DNC~\citep{Poliak_2018}, representing two recently released large-scale entailment datasets. Our choice of biLSTMs was motivated by their strong performance in recent studies~\citep{Balazs_2017, Conneau_2017}.

\texttt{Word2vec}, {\APT}s and \texttt{fastText} follow the \emph{one representation per word} paradigm~\citep{Kober_2017}, where every lexeme is represented by one vector, and contextualisation is typically achieved through distributional composition. ELMo, BERT and the pre-trained biLSTMs, on the other hand ,create context-sensitive representations on the token level. This results in different representations for the same word, depending on its current context.

\section{Experiments}
\label{experiments}
We created two experiments to assess the extent of morphosyntactic information relating to tense and aspect that is encoded in the respective embedding spaces. Subsequently we propose a novel entailment dataset and evaluate the capability of the embedding models and the pre-trained biLSTMs to perform inference on temporal predications. All our resources are available from \url{https://github.com/tttthomasssss/iwcs2019}.

\subsection{Auxiliary-Verb Agreement}
The first experiment evaluates whether the models are able to capture the agreement between an inflected verb and its corresponding aspectual auxiliary. For example, the models should be able to determine that \emph{will visit} represents a correct combination whereas \emph{will visiting} does not. We consider capturing the morphosyntactic interplay between an inflected verb and its aspectual auxiliary a pre-requisite for adequately modelling the semantics of tense and aspect.

We cast the problem as a classification task with the goal of distinguishing correct auxiliary-verb pairs from incorrect ones with a diagnostic classifier. This methodology is similar to the approach of~\citet{Linzen_2016} who assessed the ability of LSTMs to learn number agreement in English subject-verb phrases. For the dataset, we extracted verbs from the One Billion Word Benchmark (OBWB)~\citep{Chelba_2013} where each inflected verb form occurred at least 50 times. We then paired the inflected verb forms with their corresponding auxiliaries to form positive pairs, and subsequently paired each of the different inflected verb forms with all incorrect auxiliaries to build the negative pairs. We filtered the negative pairs for plausible combinations such as \emph{is eaten} by removing valid passive constructions and any invalid combination that occurred at least 5 times in the OBWB corpus. The final dataset consists of almost 36$k$ auxiliary-verb combinations with a positive : negative class distribution of 38 : 62.

\subsection{Translation Operation}
In the second experiment we assess whether it is possible to learn a translation operation between different tenses in the embedding space. We consider learning a translation operation in two ways: firstly a simple vector offset on the basis of the averaged difference between inflected verbs with their auxiliaries and their respective lemmas. Secondly, we train a feedforward neural network to project the infinitive representation of a verb to one of its inflected forms. The goal for both approaches is then to generate an unseen inflected verb form from a given unseen lemma.

The averaged offset translation is shown in Equation~\ref{eqn:offset}, where the offset $o_{t}$ is calculated on the basis of  a set of seed verbs $S$ of size $n$, and vector representations $x_{t}$ and $x_{\ell}$ of the inflected form, or contextualised form if the tense requires an auxiliary, and lemma form of the verb $x$, respectively. At prediction time, we are trying to create $x^{\prime}_{t}$ by adding the offset $o_{t}$ to the lemma $x^{\prime}_{\ell}$ (where $x^{\prime} \not\in S$). Equation~\ref{eqn:nn} shows the setup where we use a neural network to learn a translation matrix from infinitive forms to inflected forms, where $f$ is a tense-specific neural network with a single hidden layer, that takes an unseen lemma representation $x^{\prime}_{\ell}$ as input and generates an inflected form $x^{\prime}_t$, and where $\Theta_{t}$ represent the learnable parameters of the network. 
 \begin{multicols}{2}
	\noindent
	\begin{equation}
		\label{eqn:offset}
		o_{t} = \frac{1}{n}\sum_{x \in S}{x_{t} - x_{\ell}}
	\end{equation}
	\begin{equation}
		\label{eqn:nn}
		x^{\prime}_{t} = f(x^{\prime}_{\ell}; \Theta_{t})
	\end{equation}
\end{multicols}
We subsequently evaluate whether the correctly inflected verb is in the nearest neighbour list of the generated verb. The inflected verb generation setup is inspired by~\citet{Bolukbasi_2016} and~\citet{Shoemark_2017}, who used a similar method in their respective works. For the dataset, we extracted verbs from the OBWB corpus where each inflected verb form occurred at least 50 times, resulting in $\approx$2.8$k$ verbs per tense. 

\subsection{Entailment with Temporal Predications}

Lastly, we propose \textbf{TEA} --- the \textbf{T}emporal \textbf{E}ntailment \textbf{A}ssessment dataset. \textbf{TEA} contains pairs of short sentences with the same argument structure that differ in tense and aspect of the main verb, and follows a binary label annotation scheme (\emph{entailment} vs. \emph{non-entailment}). Example sentences from \textbf{TEA} are shown in Table~\ref{tbl:tea_examples}. 
\begin{table}[!htb]
\centering
\small
\resizebox{0.5\columnwidth}{!}{
\begin{tabular}{ l c l}
John \emph{is visiting} London.	& $\models$ 		& John \emph{has arrived} in London. \\
John \emph{will visit} London. 	& $\not\models$ 	& John \emph{has arrived} in London. \\\hline

John \emph{is visiting} London.	& $\not\models$	& John \emph{has left} London. \\
John \emph{is visiting} London. & $\models$ 		& John \emph{will leave} London. \\\hline

George \emph{has acquired} the house.	& $\models$		&  George \emph{owns} the house. \\
George \emph{is acquiring} the house. 	& $\not\models$ 	&George \emph{owns} the house. \\
\end{tabular}}
\captionsetup{font=small}
\caption{Examples from \textbf{TEA}.} 
\label{tbl:tea_examples}
\end{table}
The absence and infeasibility of creating a lexical resource for consequent state inference patterns creates the necessity for NLP systems to learn these rules from data. With \textbf{TEA}, we cast the problem of determining when a new consequent state is licensed by an event as a natural language inference task, thereby providing a first evaluation set for modern NLP models.

\textbf{Data Collection}. We sampled candidate pairs from the before-after category of VerbOcean~\citep{Chklovski_2004}, the WordNet verb entailment graph~\citep{Fellbaum_1998}, the entailment datasets of~\citet{Weisman_2012} and \citet{Vulic_2017}, and the relation inference dataset of~\citet{Levy_2016}. Subsequently, we manually filtered the list, and discarded candidate verb pairs without any temporal relation to each other. For each pair we chose nouns as arguments to form full sentences. The arguments further served the purpose of reducing ambiguity and avoiding habitual readings. 

\textbf{TEA} covers entailments between an all-by-all combination of the present simple, present progressive, present perfect, past simple, past progressive, past perfect and the modal future, covering perfect and progressive aspect. The dataset contains 11138 sentence pairs with a class distribution of 22 : 78 (entailment : non-entailment). More detailed dataset statistics are presented in Appendix~\ref{sec:supplemental_a}.

\textbf{Data Annotation}. We interpreted entailment as common-sense inference~\citep{Dagan_2006}, and considered a positive entailment relation between two temporal predications if a human annotator would decide that sentence 2 is \emph{most likely} true given sentence 1. We decided against a crowdsourced annotation of \textbf{TEA} as our aim was to maximise the consistency of fine-grained entailment decisions. Therefore, \textbf{TEA} was labelled by two annotators\footnote{The first and second author of this paper.}, where the first round of annotation resulted in just under 20\% disagreement across the whole dataset. The relatively high level of disagreement suggests that even for annotators who (more or less) know what they are looking for, assessing whether an entailment holds between two temporal predications is a very challenging task.


Disagreements in \textbf{TEA} were resolved on a case-by-case basis and all sentence pairs with an initial disagreement have been resolved and included in the dataset. We found that with temporality involved, suddenly \emph{everything} appeared to become uncertain. Hence we approached the disagreement resolution by first discussing which of several possible readings is the strongest, and whether that reading is sufficiently more likely than any other possible reading. Subsequently we discussed whether the strong reading is above the \emph{almost always true} threshold.

Often, disagreements resulted from different assumptions regarding the ordering of the events' nuclei. For example, even if we accept that \emph{buys} entails \emph{chooses}, \emph{will buy} does not necessarily entail \emph{will choose}. The reason is that this pair is ambiguous between two readings, a ``has-just-chosen-and-now-will-buy" reading on one hand, and a ``will-choose-and-then-will-buy" reading on the other, which seem to be equally likely in the absence of any further context\footnote{In this case we decided that if \emph{will buy} is true, the choosing didn't happen yet, so \emph{will buy} $\models$ \emph{will choose}.}. 

Even when ordering was clear, however, disagreements could arise over beliefs of when an utterance becomes licensed. Saying \emph{will graduate}, for example, can be considered reasonable at any time, or only once \emph{graduation} is sufficiently imminent and likely. In the latter case, \emph{is studying} can be considered sufficiently likely to be an entailment, while in the former case the entailment is less clear\footnote{We decided \emph{will graduate} $\models$ \emph{is studying}.}. Overall, world knowledge and intuition played into disagreements heavily, causing cases to fall just above or below the common-sense inference threshold depending on the annotator. 

We identified a possible annotation artefact in \textbf{TEA} due to our decision to annotate the dataset sequentially rather than randomly. While this greatly reduced the cognitive load, we were confronted with possible contradictions between different tenses of entailed predicates (for example, a single event cannot happen in the past \emph{and} the future). This initially led to more conservative annotations, since some pairs when viewed independently can sound very plausible. We tried to factor out this source of bias when resolving the disagreements, and are confident that the annotations in \textbf{TEA} are robust.

An interesting avenue for future work would be adding temporal adverbials to further reduce ambiguity for annotators --- and to analyse whether models can handle them correctly. The addition of temporal adverbials might alleviate the temporal ordering ambiguity, as for example reading \textit{will buy in 5 years} might help us conclude the ordering with \textit{will choose}, since \textit{choosing} is probably near \textit{buying}.

\section{Results and Analysis}
\label{results_and_analysis}
For our experiments we used the publicly available versions of each embedding model. For the evaluation on \textbf{TEA}, we trained two biLSTMs on SNLI and DNC in addition to the embedding models, achieving 83\% and 88\% accuracy on the SNLI and DNC development sets, respectively. Appendix~\ref{sec:supplemental_b} lists further details for all models.

\subsection{Auxiliary-Verb Agreement}

For assessing whether the auxiliary-verb agreement can be detected with a diagnostic classifier, we built a binary classification task, using stratified J-K-fold cross-validation~\citep{Moss_2018} and report averaged accuracy. We used the scikit-learn~\citep{Pedregosa_2011} logistic regression classifier with default hyperparameter settings.

The results in Table~\ref{tbl:agreement_results} show that the representations of {\APT}s and BERT are specific enough for a linear classifier to distinguish plausible from implausible combinations. The reason for the strong performance of {\APT}s stems from its sparsity --- plausible auxiliary-verb combinations result in representations with numerous non-zero entries, whereas implausible combinations rarely contain more than a handful of non-zero elements.
\begin{table}[!htb]
\centering
\small
\resizebox{0.8\columnwidth}{!}{
\begin{tabular}{ r | c c c c c | c}
\textbf{Auxiliary}	& \textbf{\texttt{word2vec}}	& \textbf{\APT}			& \textbf{\texttt{fastText}}	& \textbf{ELMo}		& \textbf{BERT}			& \textbf{Majority Class}	\\\hline 
is 				& 0.65 (+/- 0.02)			& 0.88 (+/- 0.01)		& 0.67 (+/- 0.02)		& 0.52 (+/- 0.01)	& \textbf{0.90} (+/- 0.01)	& 0.53 				\\
will 				& 0.48 (+/- 0.01)			& \textbf{0.94} (+/- 0.01)	& 0.58 (+/- 0.01)		& 0.63 (+/- 0.01)	& 0.89 (+/- 0.01)		& 0.67 				\\
has 				& 0.84 (+/- 0.01)			& \textbf{0.94} (+/- 0.00)	& 0.77 (+/- 0.01)		& 0.63 (+/- 0.01)	& 0.91 (+/- 0.01)		& 0.66 				\\
had 				& 0.84 (+/- 0.01)			& \textbf{0.95} (+/- 0.00)	& 0.78 (+/- 0.01)		& 0.62 (+/- 0.01)	& 0.93 (+/- 0.01)		& 0.66 				\\
was 				& 0.72 (+/- 0.02)			& 0.86 (+/- 0.01)		& 0.74 (+/- 0.02)		& 0.52 (+/- 0.01)	& \textbf{0.92} (+/- 0.01)	& 0.53 				\\\hline
\textbf{Average} 	& 0.71 (+/- 0.01)			& \textbf{0.92} (+/- 0.00)	& 0.71 (+/- 0.01)		& 0.59 (+/- 0.00)	& 0.91 (+/- 0.00)		& 0.61 				\\
\end{tabular}}
\captionsetup{font=small}
\caption{Auxiliary-verb agreeement results. Results are averaged accuracies with standard deviations in brackets.}
\label{tbl:agreement_results}
\end{table}
While \texttt{word2vec} and \texttt{fastText} seem to capture the morphosyntactic relation between an auxiliary and an inflected verb to some extent, their performance is substantially worse than {\APT}s and BERT. Somewhat surprisingly, the results for ELMo are worse than the majority class baseline for all auxiliaries. One possible reason for the comparatively weak performance of \texttt{word2vec}, \texttt{fastText} and especially ELMo in comparison to BERT is the latter's more global training objective that does not rely on sequential input. For ELMo, we also tried running it with full sentence contexts for all auxiliary-verb combinations, which, however, did not lead to improved performance (results omitted).

\subsection{Translation Operation}
For obtaining an averaged vector offset, we randomly sampled a seed set of verb types from our dataset to learn an offset vector, and subsequently aimed to predict the inflected form for all remaining verb types in the dataset. We sampled 10 different seed sets of size 10 for our experiments\footnote{In preliminary experiments we found that a seed set of 5-10 verbs is sufficient.}. 

For learning a translation operation with a neural network we used a simple feedforward architecture with a single hidden layer and a tanh activation function, using Adam with a learning rate of 0.01 to optimise the mean squared error between the generated inflected verb and the true inflected verb. Due to the neural network requiring more training data than the averaged vector offset approach, we evaluated the model using 10-fold cross-validation. For {\APT}s we projected the explicit co-occurrence space down to 100 dimensions using SVD before feeding the representations to the neural network.

Performance for both approaches is reported in terms of \emph{Mean Reciprocal Rank (MRR)}, averaged over the 10 randomly sampled seed sets and the 10 cross-validation folds, for the averaged offset vector and neural network approaches, respectively. For calculating MRR, the query space for retrieving an inflected verb, given its lemma and the computed translation operation, is based on all contextualised auxiliary-verb combinations, and all inflected forms of all verbs.

\begin{figure}[!htb]
\centering
\setlength{\belowcaptionskip}{-5pt}
\setlength{\abovecaptionskip}{-2pt}
\includegraphics[width=0.5\columnwidth]{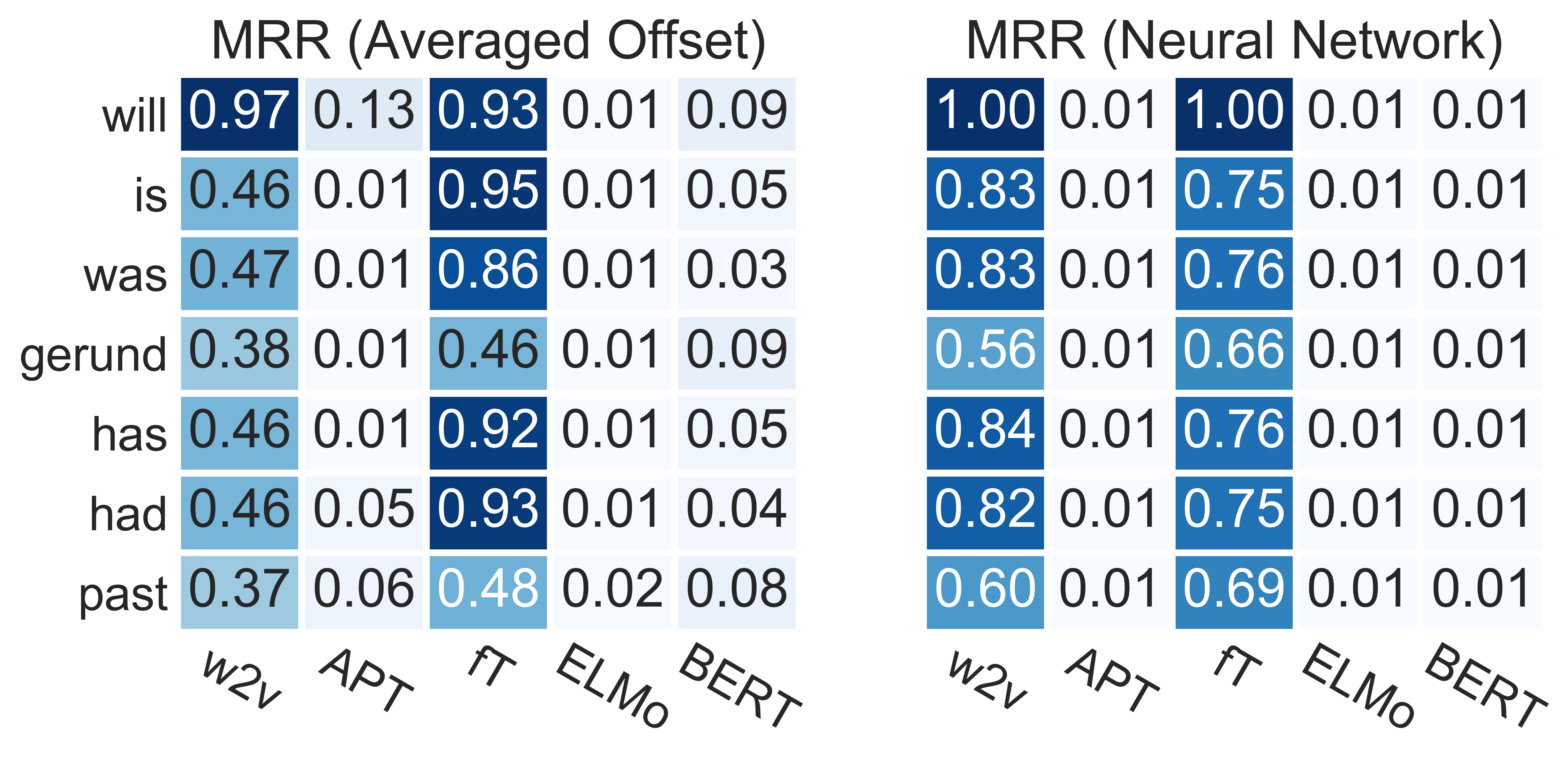}
\captionsetup{font=small}
\caption{Translation operation results based on averaged MRR.}
\label{fig:inflection_prediction_mrr}
\end{figure}

\begin{figure}[!htb]
\centering
\setlength{\belowcaptionskip}{-5pt}
\setlength{\abovecaptionskip}{-2pt}
\includegraphics[width=0.75\columnwidth]{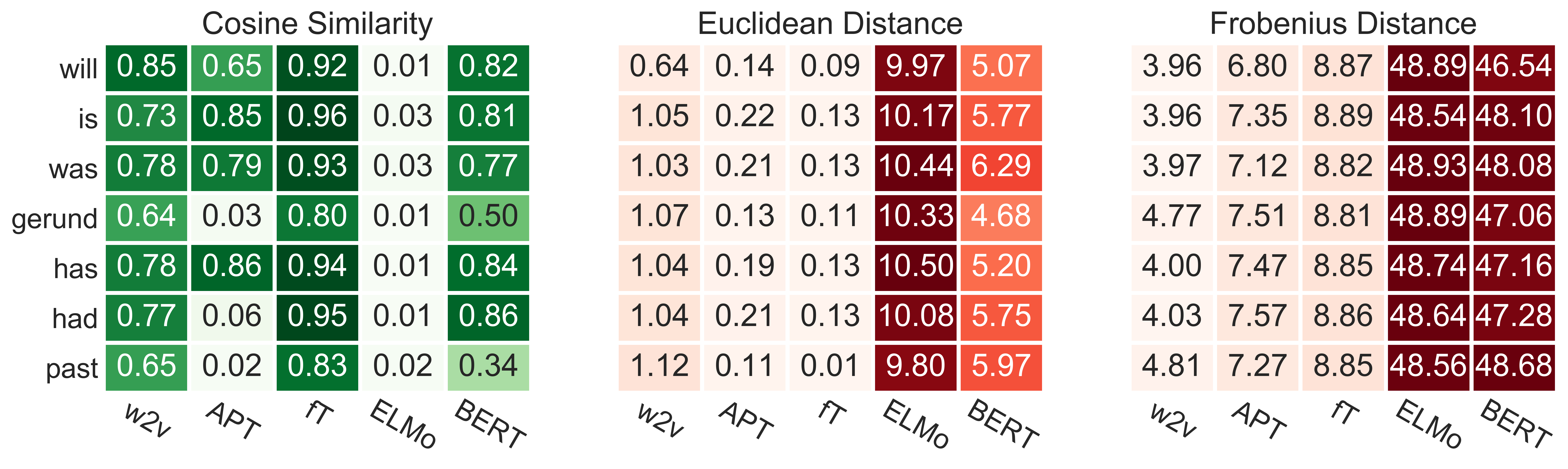}
\captionsetup{font=small}
\caption{Average cosine similarities and Euclidean distances of averaged offset vectors and Frobenius distances of the learnt neural network weight matrices.}
\label{fig:inflection_prediction_cos_euc}
\end{figure}

Creating translation operations in embedding space is primarily a word-type level task and thus potentially puts BERT and ELMo at a disadvantage as they produce representations on the token level. This is reflected in Figure~\ref{fig:inflection_prediction_mrr}, where both ELMo and BERT perform poorly in comparison to \texttt{word2vec} and \texttt{fastText}. {\APT}s also exhibit weak performance on this task, with this time the sparsity of its high-dimensional representations being disadvantageous. Interestingly, performance generally dropped --- except for \texttt{word2vec} --- when moving from the simple vector offset approach to a neural network based translation operation, providing evidence that the morphosyntax of tense and aspect is well represented as a linear offset in the embedding space.
One of the main reasons for the poor performance of ELMo and BERT was that the obtained offset vectors and learnt translation matrices varied substantially across runs. Figure~\ref{fig:inflection_prediction_cos_euc} shows the average cosine similarities (left) and average Euclidean distances (middle) between the computed offset vectors for each subtask across all 10 runs. Figure~\ref{fig:inflection_prediction_cos_euc} furthermore shows the average Frobenius distances (right) between the learnt neural network translation matrices across all 10 folds.
Figure~\ref{fig:inflection_prediction_cos_euc} mirrors the general performance trend in Figure~\ref{fig:inflection_prediction_mrr}, with vector offsets obtained from \texttt{word2vec} and \texttt{fastText} having high average cosine similarity and low average Euclidean distance. Furthermore, the lower average Frobenius distance for \texttt{word2vec} is reflected in its improved performance in comparison to \texttt{fastText} whose translation matrices exhibit a larger average Frobenius distance. For ELMo in particular, the offset vectors and translation matrices differ considerably across experimental runs. The large average Frobenius distances for ELMo and BERT also suggest that the neural network struggled to find a good minimum during learning.

\subsection{Entailment with Temporal Predications}

The results in this section so far have shown that morphosyntactic information relating to tense and aspect is encoded in the different embedding spaces. In the following we use \textbf{TEA} to analyse whether these models are able to use that information for natural language inference. As our goal is to assess to what extent tense and aspect are captured by the models, we refrain from fine-tuning them on \textbf{TEA}.

For evaluation we measure precision and recall over varying thresholds and report performance in terms of average precision\footnote{Also known as the area under the precision-recall curve.}. \textbf{TEA} can also serve as an additional evaluation set for sentence encoder models trained on large-scale natural language inference datasets such as SNLI or DNC, which themselves include very little temporal information in their respective test sets. We therefore additionally cast \textbf{TEA} as a binary classification task, and report accuracy and macro-averaged F1-score for the two pre-trained biLSTM models.

Table~\ref{tbl:tea_results} shows the average precision scores for the models and the accuracy and F1-scores for the two pre-trained biLSTMs in comparison to a majority class baseline and a baseline predicting the majority class per tense pair. We used cosine as similarity measure for the embedding models and the softmax prediction scores for the biLSTMs. For {\APT}s, we also tried the asymmetric inclusion score BInc~\citep{Szpektor_2008}, however found cosine working better. We furthermore experimented with distributional inference~\citep{Kober_2016}, and found a small positive impact on recall but a slightly larger negative dip in precision, which overall led to slightly lower average precision scores.
\begin{table}[!htb]
\centering
\small
\resizebox{0.5\columnwidth}{!}{
\begin{tabular}{ l | c c c }
\textbf{Model} 			& \textbf{Avg. Precision}	& \textbf{Accuracy}	& \textbf{F1-Score} \\\hline
\texttt{word2vec}		& 0.31				& -				& - \\
\APT					& 0.28				& -				& - \\
\texttt{fastText}			& 0.30				& -				& - \\
ELMo				& 0.21				& -				& - \\
BERT				& 0.27				& -				& - \\
biLSTM-DNC			& 0.22 				& 0.58 			& 0.49 \\
biLSTM-SNLI			& 0.21 				& 0.51 			& 0.47 \\\hline
Maj. class				& 0.22				& 0.78			& 0.44 \\
Maj. class / tense pair	& \textbf{0.35} 			& \textbf{0.80} 		& \textbf{0.66}\\\hline
\end{tabular}}
\captionsetup{font=small}
\caption{\textbf{TEA} results. All model results are significantly worse at the $p < 0.01$ level w.r.t. the majority class / tense pair baseline, using a randomised bootstrap test~\citep{Efron_1994}.}
\label{tbl:tea_results}
\end{table}

The results show that neither of the models are able to outperform the majority class / tense baseline. This highlights that despite the use of short and simple sentences in the dataset, the latent nature of tense and aspect make \textbf{TEA} a very challenging problem.

\begin{figure}[!htb]
\centering
\setlength{\belowcaptionskip}{-5pt}
\setlength{\abovecaptionskip}{-2pt}
\includegraphics[width=0.55\columnwidth]{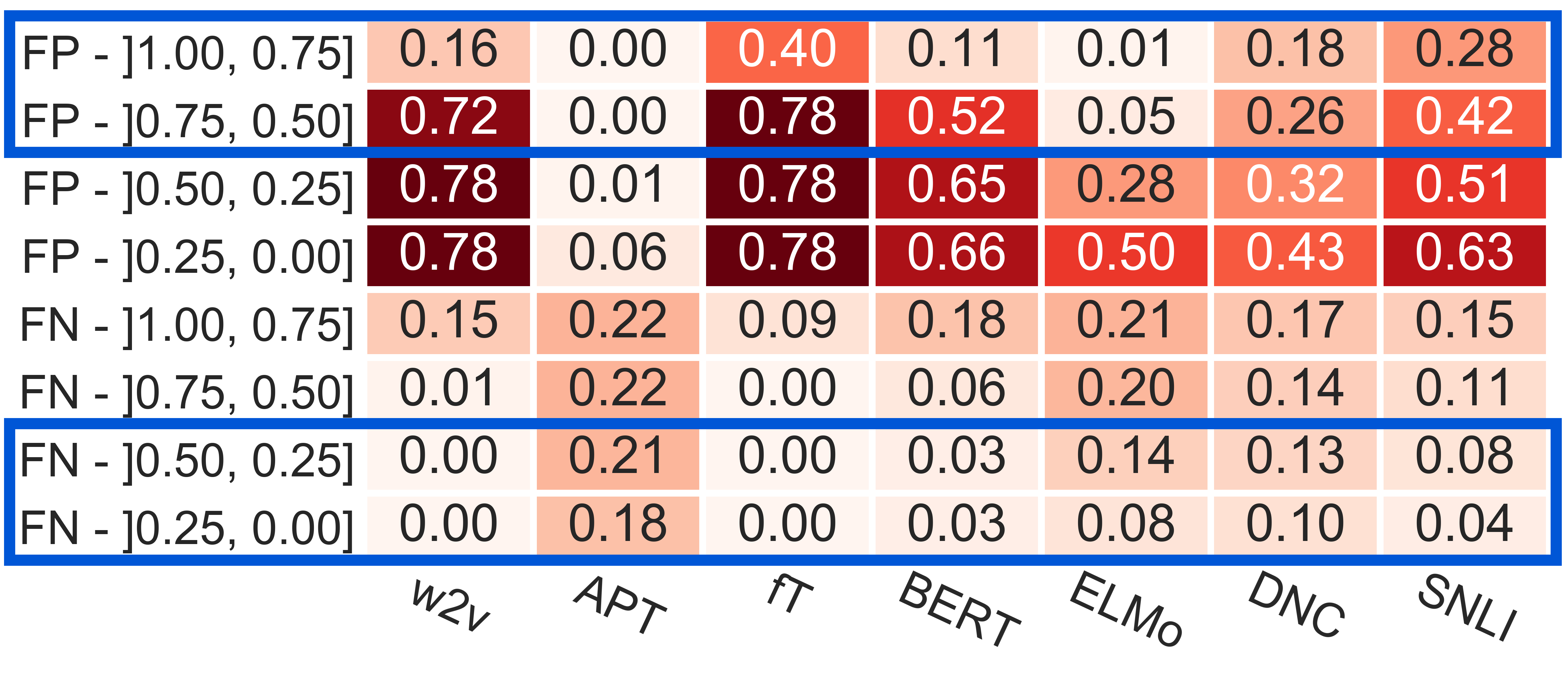}
\captionsetup{font=small}
\caption{False Positive (FP) and False Negative (FN) rates.}
\label{fig:fp_fn_rates}
\end{figure}

In order to analyse the causes for the low performance across models, we calculated the false positive and false negative rates for different similarity threshold ranges for each of the models. Figure~\ref{fig:fp_fn_rates} shows that even for high thresholds, the neural embedding models frequently predict entailment when there isn't one, thereby producing a high rate of false positives (highlighted at the top of Figure~\ref{fig:fp_fn_rates}). Conversely, a sparse model such as {\APT}s, fails to predict entailment when there actually is one, resulting in a high rate of false negatives (highlighted at the bottom of Figure~\ref{fig:fp_fn_rates}). 
Our results show that natural language inference on temporal predications is a challenging problem, especially for distributional semantic approaches. One reason is that these models are primarily governed by contextual similarity which is a bad proxy for inference in the case of a dataset such as \textbf{TEA}. For example, if \emph{Jane has arrived in London}, then she \emph{was going to London} at some earlier point, but it is not the case that she currently \emph{is going to London}. Furthermore, when she \emph{has arrived in London}, she \emph{is visiting London} at the moment, and \emph{will leave} again at some point in the future. 

The predications in the short narrative above are very diverse in terms of tense and aspect, however the main verbs --- or even the predications as a whole --- typically have high distributional similarity, which inevitably leads to numerous false entailment decisions as reflected in Figure~\ref{fig:fp_fn_rates}.

In the following we briefly analyse the impact of distributional similarity and investigate to what extent the similarity scores between two predications change when tense and aspect influence the entailment. Table~\ref{tbl:qualitative_analysis} shows that the cosine similarity between temporally and aspectually modified predications is typically higher than for their respective lemmas. This further indicates that many false positives of the neural network based models in our results are due to high distributional similarity scores between predications. For {\APT}s the cosine scores --- even when normalised --- are generally very low due to their sparsity and high dimensionality, highlighting their bias towards false negatives.
\begin{table}[!htb]
\centering
\small
\resizebox{0.6\columnwidth}{!}{
\begin{tabular}{ r | c c c c c c c}
\textbf{Predication Pair} 			& \textbf{\texttt{w2v}}	& \textbf{\APT}	& \textbf{\texttt{fT}}	& \textbf{ELMo}		& \textbf{BERT}		& \textbf{DNC}	& \textbf{SNLI}	\\\hline 
visit $\models$ leave				& 0.36 			& 0.09		& 0.53			& 0.59 			& 0.69 			& 0.69		& 0.28		\\
is visiting $\models$ will leave 		& 0.57  			& 0.02 		& 0.60			& 0.60 			& \textbf{0.77} 		& 0.26		& \textbf{0.26}	\\
is visiting $\not\models$ has left 	& 0.58  			& 0.03 		& 0.71			& 0.65 			& 0.72 			& 0.32		& 0.20		\\\hline
visit $\models$ arrive 			& 0.45  			& 0.07		& 0.55			& 0.49 			& 0.71 			& 0.58		& 0.45		\\
is visiting $\models$ has arrived 	& \textbf{0.62} 		& \textbf{0.04}	& \textbf{0.69}		& \textbf{0.51} 		& \textbf{0.84} 		& 0.25		& \textbf{0.51}	 \\
is visiting $\not\models$ will arrive 	& 0.57 			& 0.01		& 0.60			& 0.50 			& 0.81 			& 0.32		& 0.25		\\\hline
win $\models$ play 				& 0.52			& 0.14 		& 0.54  			& 0.59 			& 0.73 			& 0.39		& 0.32		\\
has won $\models$ has played 		& \textbf{0.75}		& \textbf{0.25} 	& \textbf{0.88} 		& \textbf{0.60} 		& \textbf{0.85} 		& \textbf{0.55}	& 0.23		\\
has won $\not\models$ will play 	& 0.60  			& 0.11 		& 0.64			& 0.55			& 0.78			& 0.31		& 0.36		\\\hline
\end{tabular}}
\captionsetup{font=small}
\caption{Similarity scores between the example predicates. DNC and SNLI refer to the two biLSTMs pre-trained on DNC and SNLI, respectively.}
\label{tbl:qualitative_analysis}
\end{table}

However, Table~\ref{tbl:qualitative_analysis} also shows that in most cases the distributional similarity between an entailed pair is higher than for a non-entailed pair (boldfaced in Table~\ref{tbl:qualitative_analysis}). This indicates that the embedding models do appear to capture \emph{some} of the semantics of tense and aspect in their respective contextualised representations. However, their high distributional similarity overwhelms any finer distinction that the models might have extracted.

While our analysis indicates that the embedding models are able to extract knowledge about tense and aspect, the signal is not strong enough to reliably perform inference. A potential avenue for future work would therefore be the development of models that are able to better represent tense and aspect, while not being primarily governed by distributional similarity. 

\section{Related Work}
\label{related_work}
Most previous work on inference between verbs was concerned with extracting inference rules from raw text~\citep{Lin_2001,Szpektor_2004,Szpektor_2007,Hashimoto_2009,Melamud_2013a}. As a next step,~\citet{Berant_2010} and~\citet{Hosseini_2018} leverage these rules to build entailment graphs for modelling natural language inference. However in both cases the entailment graphs are built on the basis of \emph{verb lemmas} and do not take tense and aspect into account. One example of using tense for inference is~\citet{Pavlick_2016b}, who leverage implicative verbs to determine that \emph{managed to solve X} $\models$ \emph{X is solved}.
Our proposed dataset \textbf{TEA} fills a gap in the natural language inference evaluation repertoire by focusing on temporal and aspectual entailment. Recent years saw the release of a number of large-scale datasets, such as SNLI~\citep{Bowman_2015}, MNLI~\citep{Williams_2017} or DNC~\citep{Poliak_2018}, but neither of these datasets focuses on, or includes a substantial proportion of, inference examples between temporal predications. 

\textbf{TEA} is related to work on causality~\citep{Mirza_2014,Mirza_2014b}, however our dataset has been created from scratch rather than derived from TimeBank~\citep{Pustejovsky_2003}, as for example explicit \emph{buys} $\models$ \emph{owns} relations are rarely encountered in the same paragraph or connected by explicit causal links. Therefore, \textbf{TEA} captures many consequent state inferences that are missing from previous datasets. The most closely related task to \textbf{TEA} is the relation inference dataset of~\citet{Levy_2016}, which however, contains only very few examples where temporality is a governing factor.







\section{Future Work}
\label{future_work}
In future work we plan to leverage tense- and aspect-based information for constructing temporal entailment graphs~\citep{Lewis_2014}, where nodes represent tensed predicates (e.g. \emph{has visited}), and edges represent entailment relations. Temporal entailment graphs, together with knowledge about the \emph{completedness} or \emph{current relevance} of an event, can be applied to procedural reasoning, such as tracking the state of entities through text, similar to recent work of~\citet{Bosselut_2017}, and~\citet{Henaff_2017}. We furthermore plan to focus on other types of aspect such as \emph{Aktionsart}.

\section{Conclusion}
In this paper we highlighted that tense and aspect are two of the most important factors for performing natural language inference. We introduced a novel entailment dataset, \textbf{TEA}, that contains pairs of short sentences and focuses on entailment relations between temporally and aspectually modified verbs. We showed that distributional embedding models capture a considerable amount of the morphosyntactic information relating to tense and aspect in their embedding spaces. However, neither the embedding models, nor two pre-trained biLSTMs, were able to outperform a simple rule-based baseline on \textbf{TEA}, primarily due to their reliance on contextual similarity for inference. In this sense, tense and aspect semantically resemble logical operators like negation rather than distributional components. The challenge will be to combine logical operator semantics with distributional representations of content words.
\label{conclusion}

\section*{Acknowledgements}
We thank Javad Hosseini, Paola Merlo and Nate Chambers for valuable discussions and comments on this work. We also thank our anonymous reviewers for their helpful feedback which led to a substantially improved paper. This research was supported in part by ERC Advanced Fellowship GA 742137 SEMANTAX, a Google faculty award, a Bloomberg L.P. Gift award, and a University of Edinburgh/Huawei Technologies award to Mark Steedman.
\label{acknowledgements}

\bibliographystyle{chicago}
\bibliography{iwcs2019}

\clearpage
\appendix

\section{Supplemental Material}
\label{sec:supplemental_a}
\subsection{Dataset Details}

Table~\ref{tbl:tea_statistics} shows a detailed overview of the number of examples per tense and aspect pair, as well as their class distribution.
\begin{table}[!htb]
\centering
\small
\resizebox{0.8\columnwidth}{!}{
\begin{tabular}{ l | c | c}
\textbf{Category} 						& \textbf{Num. Examples}	& \textbf{Class distribution} (\emph{entailment : non-entailment}) \\\hline
Present progressive - Present progressive		& 188				& 33 : 67 \\
Present progressive - Past progressive 		& 188				& 23 : 77 \\
Present progressive - Present perfect 		& 213 				& 20 : 80 \\
Present progressive - Past perfect 			& 213 				& 12 : 88 \\
Present progressive - Future simple 			& 216 				& 28 : 72 \\
Present progressive - Present simple 		& 216 				& 27 : 73 \\
Present progressive - Past simple 			& 216 				& 26 : 74 \\

Past progressive - Present progressive 		& 188 				& 0 : 100 \\
Past progressive - Past progressive 			& 188 				& 55 : 45 \\
Past progressive - Present perfect 			& 213 				& 7 : 93 \\
Past progressive - Past perfect 				& 213 				& 46 : 54 \\
Past progressive - Future simple 			& 216 				& 1 : 99 \\
Past progressive - Present simple 			& 216 				& 0 : 100 \\
Past progressive - Past simple 				& 216 				& 49 : 51 \\

Present perfect - Present progressive 		& 213 				& 12 : 88 \\
Present perfect - Past progressive 			& 213 				& 44 : 56 \\
Present perfect - Present perfect 			& 240 				& 26 : 74 \\
Present perfect - Past perfect 				& 240 				& 26 : 74 \\
Present perfect - Future simple 				& 243 				& 16 : 84 \\
Present perfect - Present simple 			& 243 				& 17 : 83 \\
Present perfect - Past simple 				& 243 				& 42 : 58 \\

Past perfect - Present progressive 			& 213 				& 0 : 100 \\
Past perfect - Past progressive 				& 213 				& 58 : 42  \\
Past perfect - Present perfect 				& 240 				& 3 : 97 \\
Past perfect - Past perfect 				& 240 				& 59 : 41 \\
Past perfect - Future simple 				& 243 				& 0 : 100  \\
Past perfect - Present simple 				& 243 				& 0 : 100 \\
Past perfect - Past simple 					& 243 				& 58 : 42 \\

Future simple - Present progressive 			& 216 				& 3 : 97 \\
Future simple - Past progressive 			& 216 				& 1 : 99 \\
Future simple - Present perfect 				& 243 				& 1 : 99 \\
Future simple - Past perfect 				& 243 				& 1 : 99 \\
Future simple - Future simple 				& 246 				& 47 : 53 \\
Future simple - Present simple 				& 246 				& 2 : 98 \\
Future simple - Past simple 				& 246 				& 1 : 99 \\

Present simple - Present progressive 		& 216 				& 21 : 79 \\
Present simple - Past progressive 			& 216 				& 29 : 71 \\
Present simple - Present perfect 			& 243 				& 15 : 85 \\
Present simple - Past perfect 				& 243 				& 17 : 83 \\
Present simple - Future simple 				& 246 				& 19 : 81 \\
Present simple - Present simple 			& 246 				& 29 : 71 \\
Present simple - Past simple 				& 246 				& 26 : 74 \\

Past simple - Present progressive 			& 216 				& 0 : 100 \\
Past simple - Past progressive 				& 216 				& 55 : 45 \\
Past simple - Present perfect 				& 243 				& 5 : 95 \\
Past simple - Past perfect 					& 243 				& 54 : 46 \\
Past simple - Future simple 				& 246 				& 0 : 100 \\
Past simple - Present simple 				& 246 				& 1 : 99 \\
Past simple - Past simple 					& 246 				& 56 : 44 \\\hline

Progressive - Progressive					& 3464				& 20 : 80 \\
Progressive - Perfect						& 2748				& 18 : 82 \\
Perfect - Progressive						& 2748				& 16 : 84 \\
Perfect - Perfect						& 2178				& 37 : 63 \\\hline\hline

\emph{TOTAL}							& 11138				& 22 : 78\\
\end{tabular}}
\captionsetup{font=small}
\caption{Detailed statistics of \textbf{TEA}.} 
\label{tbl:tea_statistics}
\end{table}

\clearpage
\section{Supplemental Material}
\label{sec:supplemental_b}
\subsection{Model Details}

\textbf{\texttt{word2vec}.} We used the 300-dimensional vectors trained on GoogleNews, available from \url{https://code.google.com/archive/p/word2vec/}.

\noindent\textbf{{\APT}s.} We used order 2 {\APT}s trained on Gigaword, with PPMI weighting and no negative SPPMI shift which were used in~\citet{Kober_2018}. As composition function we used \emph{composition by intersection} which has previously been shown to work well for modelling the similarity of short phrases~\citep{Kober_2016,Kober_2017b}.

\noindent\textbf{\texttt{fastText}.} We used the 300-dimensional pre-trained vectors with subword information trained on Wikipedia~\citep{Mikolov_2018}.

\noindent\textbf{ELMo.} We are using the pre-trained model released by~\citet{Peters_2018} and accessible via the AllenNLP toolkit~\citep{Gardner_2017}.

\noindent\textbf{BERT.} We are using the BERT-big model released by~\citet{Devlin_2018} and available from \url{https://github.com/google-research/bert}.

\noindent\textbf{Pre-trained biLSTM.} We are using a bi-directional LSTM~\citep{Hochreiter_1997} with max pooling, but without an attention layer. We follow~\citet{Balazs_2017} in aggregating the embedded and pooled premise and hypothesis representations before passing them to a single fully connected layer, with a relu activation function and a dropout~\citep{Srivastava_2014} probability of 0.3. The model is optimised with Adam~\citep{Kingma_2014} using a learning rate of 0.01. The model is implemented in PyTorch~\citep{Paszke_2017}. Table~\ref{tbl:bilstm_results} lists the accuracies on the SNLI and DNC development and test sets for our model.
\begin{table}[!htb]
\centering
\small
\resizebox{0.4\columnwidth}{!}{
\begin{tabular}{l | c | c}
\textbf{Dataset}	& \textbf{Dev Accuracy}	& \textbf{Test Accuracy}	\\\hline 
SNLI 		&  0.83				& 0.82 				\\
DNC			&  0.88				& 0.87				\\\hline
\end{tabular}}
\captionsetup{font=small}
\caption{Accuracies on the development and test sets for the pre-trained biLSTMs on SNLI and DNC, respectively.}
\label{tbl:bilstm_results}
\end{table}

\end{document}